\documentclass{article}

\usepackage{microtype}
\usepackage{graphicx}
\usepackage{subfigure}
\usepackage{booktabs} %

\usepackage{hyperref}

\usepackage[accepted]{icml2024}

\usepackage{amsmath}
\usepackage{amssymb}
\usepackage{mathtools}
\usepackage{amsthm, eucal}

\usepackage[capitalize,noabbrev]{cleveref}
\theoremstyle{plain}
\newtheorem{theorem}{Theorem}[section]

\newtheorem{lemma}[theorem]{Lemma}

\theoremstyle{definition}
\newtheorem{definition}[theorem]{Definition}

\theoremstyle{remark}
\newtheorem{remark}[theorem]{Remark}

\usepackage{colortbl}
\usepackage{xcolor}
\usepackage{hyperref}
\hypersetup{
    colorlinks=true,
    breaklinks=true,
    urlcolor=magenta,
    linkcolor=red,
    bookmarksopen=false,
    filecolor=black,
    citecolor=blue,
}

\usepackage[textsize=tiny]{todonotes}

\icmltitlerunning{Generative Active Learning for Long-tailed Instance  Segmentation}

\begin{document}

\twocolumn[
\icmltitle{Generative Active Learning for Long-tailed Instance  Segmentation}

\icmlsetsymbol{equal}{*}

\begin{icmlauthorlist}
\icmlauthor{Muzhi Zhu}{zju,equal}
\icmlauthor{Chengxiang Fan}{zju,equal}
\icmlauthor{Hao Chen}{zju}
\icmlauthor{Yang Liu}{zju}
\icmlauthor{Weian Mao}{adelade,zju}
\icmlauthor{Xiaogang Xu}{zju}
\icmlauthor{Chunhua Shen}{zju}
\end{icmlauthorlist}

\icmlaffiliation{zju}{Zhejiang University, China}
\icmlaffiliation{adelade}{The University of Adelaide, Australia}

\icmlcorrespondingauthor{Hao Chen}{haochen.cad@zju.edu.cn}
\icmlcorrespondingauthor{Chunhua Shen}{chunhuashen@zju.edu.cn}

\icmlkeywords{Machine Learning, ICML}

\vskip 0.3in
]

\printAffiliationsAndNotice{\icmlEqualContribution} %

\begin{abstract}
Recently, large-scale language-image generative models have gained widespread attention and many works have utilized generated data from these models to further enhance the performance of perception tasks.
However, not all generated data can positively impact downstream models, and these methods do not thoroughly explore how to better select and utilize generated data. 
On the other hand, there is still a lack of research oriented towards active learning on generated data.
In this paper, we explore how to perform active learning specifically for generated data in the long-tailed instance segmentation task.
Subsequently, we propose BSGAL, a new algorithm that online estimates the contribution of the generated data based on gradient cache. 
BSGAL can handle unlimited generated data and complex downstream segmentation tasks effectively.
Experiments show that BSGAL outperforms the baseline approach and effectually improves the performance of long-tailed segmentation. Our code can be found at \href{https://github.com/aim-uofa/DiverGen}{https://github.com/aim-uofa/DiverGen}.
 
\end{abstract}

\section{Introduction}
\label{submission}
Data is one of the driving forces behind the development of artificial intelligence. In the past, securing high-quality data was a time-consuming and laborious task. Yet, a large amount of high-quality data is crucial for a model to achieve breakthrough performance.  Therefore, many active learning methods have emerged to explore the most informative samples from massive unlabeled data to achieve better model performance with minimal annotation costs.
Currently, the rapid development of generative models has made it possible to obtain massive amounts of high-quality data, including long-tailed data, at a relatively low cost.
In the field of visual perception, there have been many works utilizing generated data to improve perception tasks, including classification \cite{azizi2023synthetic}, detection \cite{chen2023integrating}, and segmentation \cite{Wu_2023_ICCV}. However, they often directly use generated samples as mixed training data \cite{freemask} or as data augmentation \cite{zhao2022XPaste} without exploring how to better filter and utilize the data for downstream models.

On the other hand, existing data mining and filtering methods, such as active learning, have only been validated on real data and are not suitable for generated data, as there are differences between generated data and real data regarding characteristics and usage scenarios. 
The differences are mainly as follows:
1) Existing data analysis methods are aimed at a limited data pool, while the scale of generated data is almost infinite.
2) Active learning is often carried out under a specified annotation budget. Thanks to the development of conditional generative models, the annotation cost of generated data can be almost negligible. However, this results in an unclear and uncertain quality of annotation in generated data compared to expert annotations.
3) There are differences in the distribution between real data and generated data, while the target data of previous methods do not have obvious distribution differences. 

\begin{figure}[t]
    \centering
    \includegraphics[width=1\linewidth]{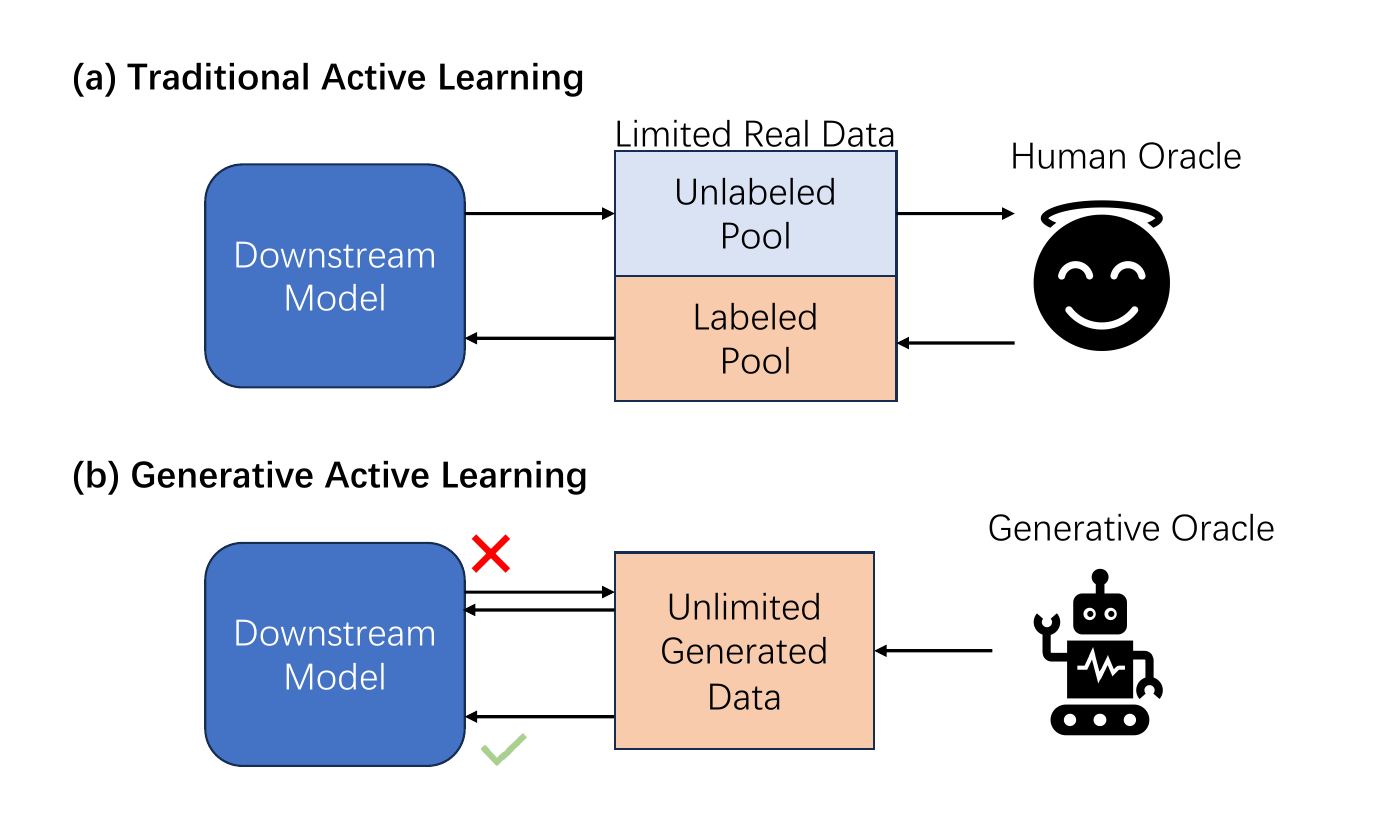}
    \vspace*{-0.5cm}
    \caption{ Comparison between Traditional Active Learning and Generative Active Learning frameworks. (a) Traditional Active Learning relies on a human oracle, therefore the annotation is accurate but with a limited budget, so the model is required to select the most informative unlabeled data. (b) Generative Active Learning, which relies on a generative oracle, has an unlimited labeled pool. However, the quality of annotation varies greatly, so the model must judiciously accept data.}
    \vspace*{-0.5cm}
    \label{fig:gal}
\end{figure}

In response to the aforementioned challenges, we propose a novel problem called ``Generative Active Learning for Long-tailed Instance Segmentation'' (see \cref{fig:gal}), which investigates how to utilize generated data effectively for downstream tasks. 
We focus on long-tailed instance segmentation for three main reasons.
First, data collection for long-tailed categories is exceedingly arduous, and how to do classification well is currently a focus in the segmentation field \cite{Kirillov_2023_ICCV,li2023semantic,yuan2024open,wei2024segment}. Given that generated data has demonstrated the potential to alleviate this difficulty \cite{zhao2022XPaste, xie2023mosaicfusion,fan2024divergen}, it is necessary to introduce generated data for long-tail segmentation tasks.
Second, the quality requirements for generated data in long-tailed instance segmentation tasks are very high, and not all generated data can have a promoting effect. Therefore, it is necessary to further explore how to screen generated data.
Third, this task itself is very comprehensive and challenging, which has more guiding significance for migration to real-world scenarios.

Here, 
our objective is to design a generative active learning pipeline targeted at long-tailed image segmentation tasks. 
Inspired by data-influence analysis methods \cite{ling1984residuals,koh2017understanding}, we first use the change of the loss value to provide an estimation of the contribution of a single generated instance in the ideal case, as discussed  in \cref{sec:ideal}. Employing the first-order Taylor expansion, we introduce an approximate contribution estimation function based on the gradient dot product, which avoids repeated calculations on the test set in the offline setting. Based on this technique, we conduct a toy experiment on CIFAR-10 \cite{krizhevsky2009learning} in \cref{sec:toy} along with a qualitative analysis of each sample on the LVIS dataset in \cref{sec:qualitative}, which preliminarily verify the feasibility of our approach.
Subsequently, in \cref{sec:batch} we explore how to apply this evaluation function to the actual segmentation training process. 
We propose the Batched Streaming Generative Active Learning algorithm (BSGAL), which allows for online acceptance or rejection of each batch of generated data. Additionally, based on the first-order gradient approximation, we maintain a gradient cache based on momentum updates to enable a more stable contribution estimation.
Finally,  experiments are carried out on the LVIS dataset, establishing that our method outperforms both unfiltered or CLIP-filtered counterparts under various backbones.  Notably, in the long-tailed category, there is an over $10\%$ improvement in $AP_r$  \cite{gupta2019lvis}. In addition,
We conduct a series of ablation experiments to delve into the particulars of our algorithmic design, including the choice of loss, the way of contribution estimation, and the sampling strategy of the test set.
To summarize, our primary contributions are detailed below:
\begin{itemize}
     \vspace*{-0.2cm}
    \item We introduce a novel problem called “Generative Active Learning for
Long-tailed Instance Segmentation”:  how to design an effective method focused on the successful using generated data, aimed at enhancing the performance of downstream segmentation tasks.  Existing data analysis methods are neither directly applicable to generated data nor have they been affirmed as efficient for such data.
    \item We propose a batched streaming generative active learning method (BSGAL) based on gradient cache to estimate generated data contribution. This pipeline can adapt to the actual batched segmentation training process, handle unlimited generative data online, and effectively enhance the performance of the model. 
    \item We carry out experiments on the LVIS dataset and demonstrate that our method outperforms both unfiltered and CLIP-filtered methods. Our model surpasses the baseline by +1.2 on $AP_{box}$ and +3.62 on $AP_r$.   We also conduct more comprehensive analysis experiments on the design or hyperparameters of our method.
\end{itemize}

\section{Related work}
\subsection{Generative Data Augmentation}
Generative data augmentation (GDA) refers to using generative models to synthesize additional data for augmentation. With the continuous improvement in the capabilities of generative models~\cite{goodfellow2020generative,saharia2022photorealistic,rombach2021highresolution}, GDA has become a popular technique for improving model performance. Several works use GDA in perceptual tasks such as classification~\cite{feng2023diverse,zhang2023prompt,azizi2023synthetic}, detection~\cite{zhao2022XPaste,chen2023integrating}, and segmentation~\cite{li2023open,Wu_2023_ICCV,wu2023datasetdm,xie2023mosaicfusion}. Early works~\cite{zhang2021datasetgan,li2022bigdatasetgan} involve the use of generative models like Generative Adversarial Networks (GANs)~\cite{goodfellow2020generative} to generate additional training data. With the evolution of diffusion models, recent works~\cite{azizi2023synthetic,li2023open,Wu_2023_ICCV,freemask,zhao2022XPaste} favor using high-quality diffusion models such as Imagen~\cite{saharia2022photorealistic} and Stable Diffusion~\cite{rombach2021highresolution} for data generation. 
X-Paste~\cite{zhao2022XPaste} has proven the strategy of using copy-paste to be more effective than directly using generated data for mixed training, and for the first time demonstrated that using generated data can enhance the performance of segmentation models on the long-tailed segmentation dataset LVIS \cite{gupta2019lvis}. Therefore, we consider it as the baseline for our work.
However, while previous works have investigated the effects of GDA on different tasks, there has been limited exploration on how to better filter and utilize generative data for downstream perception models.

\subsection{Active Learning and Data Analysis}

Analysis of the information or contribution of data samples to a model has been extensively studied long before the advent of deep learning.
Among them, two fields are most relevant to our work, one is active learning, and the other is training data influence analysis.

Active learning \cite{ren2021survey} mainly focuses on how to explore the most informative samples from massive unlabeled data to achieve better model performance with minimal annotation costs. Generally speaking, active learning can be divided into two categories. One is uncertainty-based active learning, which measures the uncertainty of samples by the posterior probability of the predicted category \cite{lewis1994heterogeneous,lewis1995sequential,goudjil2018novel} or the entropy of the predicted distribution \cite{joshi2009multi,luo2013latent}, and then selects the most uncertain samples for annotation.
The other is diversity-based active learning, which is based on clustering \cite{nguyen2004active} or core-set \cite{sener2018active} methods. They attempt to mine the most representative samples from the data to achieve minimal annotation costs.
Recently, active learning in deep learning also tends to adopt a batch-based sample querying method \cite{Ash2020Deep}, which is consistent with our work.
The most relevant work to our work is VeSSAL \cite{pmlr-v202-saran23a}, which does batched active learning in a streaming setting and samples in a gradient space. Another relatively related work \cite{mahapatra2018efficient} trains a  GAN on medical images, using the GAN to generate more data for active learning.

Training data influence analysis \cite{hammoudeh2022training} explores the relationship between training data samples and model performance, which can be divided into retraining-based \cite{ling1984residuals,roth1988shapley,feldman2020neural} and gradient-based \cite{koh2017understanding,yeh2018representer}.
The most typical retraining-based method is Leave-One-Out \cite{ling1984residuals,jia2021scalability}, which measures the contribution of a sample to the model by removing a sample from the training set and then retraining the model. However, this method is obviously impractical for modern large-scale datasets.
Therefore, many gradient-based methods have emerged recently, which use gradients to approximate the change of loss, such as using first-order Taylor expansion or Hessian matrix, to estimate the influence of samples.
The most relevant work to ours
is TracIn \cite{pruthi2020estimating}, which implements heuristic dynamic estimation through first-order gradient approximation and stored checkpoints. Unlike our work, the ultimate goal of TracIn is to estimate and filter out mislabeled samples in the training set through self-influence. 
Moreover, TracIn is only applicable to small-scale classification datasets, it is difficult to migrate to larger and complex tasks like segmentation, let alone handle nearly infinite generated data.
Our work succeeds in designing an automated pipeline for utilizing generated data to enhance downstream perception tasks.

 Most importantly, the above work is all done on relatively simple classification tasks, and only a few works have explored more complex perception tasks such as detection \cite{shrivastava2016training,liu2021influence} and segmentation \cite{jain2016active,vezhnevets2012active, Casanova2020Reinforced}, but they are all aimed at real data. 
 Our work is the first to explore the generated data on the complex perception task of long-tail instance segmentation.

\section{Preliminary} 
Formally, in generated data augmented instance segmentation tasks, we have a set of labeled real data 
$\mathcal{R}=\left\{\left(\mathbf{I}_r, \mathbf{Y}_r\right)\right\}$ and a set of generated data (with noisy label)  $\mathcal{G}=\left\{\left({\mathbf{I}}_g, {\mathbf{Y}}_g\right)\right\}$  where $\mathbf{I}$  is the image and $\mathbf{Y}$ is the label for instance segmentation. %
 Our goal is to effectively utilize the existing data $\mathcal{R}$ and $\mathcal{G}$ to train a segmentation network $f$ parameterized by $\theta$ to achieve optimal performance on unseen test data $\mathcal{U}=\left\{\left(\mathbf{I}_u, \mathbf{Y}_u\right)\right\}$, that is, minimize the loss on the test set $L_\mathcal{U}(\theta)=\sum_{(\mathbf{I}_u, \mathbf{Y}_u) \in \mathcal{U}} \ell(\mathbf{I}_u, \mathbf{Y}_u ; {\theta})$ , where $\ell$ represents the loss of the segmentation network.
The most direct way to utilize the two types of data is to perform joint training, and X-Paste proves that using the copy-paste \cite{ghiasi2021simple} method to add instances from $\mathbf{I}_g$ to real images $\mathbf{I}_r$ can achieve better results. Although our method is universal, we build it upon a copy-paste baseline.
For the convenience of subsequent description, we record this operation as $Copypaste$ such that $\widehat{\mathbf{I}}_r, \widehat{\mathbf{Y}}_r  = Copypaste(\mathcal{G}_b, \mathbf{I}_r)$, where $\mathcal{G}_b \in \mathcal{G}$ is a subset sampled from $\mathcal{G}$, $\widehat{\mathbf{I}}_r$ and $\widehat{\mathbf{Y}}_r$ represent the image and the label obtained after pasting the instance in $\mathcal{G}_b$ to $\mathbf{I}_r$.

\setlength{\textfloatsep}{10pt}
\begin{algorithm}[tb]
   \caption{Pipeline for copy-paste baseline}
   \label{alg:X-Paste}
\begin{algorithmic}[1]
    \REQUIRE labeled real data $\mathcal{R}$, generated data $\mathcal{G}$, batch size $B$, number of iterations $T$, pretrained segmentation network $f$ with parameters $\theta$, maximum number of paste instances $K$ for each image
    \FOR{$t=1$ {\bfseries to} $T$}
    \STATE Sample a batch of real data $\mathcal{R}_b \in \mathcal{R}$ 
    \FOR{$\mathbf{I}_r \in \mathcal{R}_b$}
    \STATE Get a random number $k$ from $[0, K]$
    \STATE Sample $k$ instances from $\mathcal{G}$ in a class-balanced way to get $\widehat{\mathbf{I}}_r, \widehat{\mathbf{Y}}_r  = Copypaste(\mathcal{G}_b, \mathbf{I}_r)$
    \ENDFOR
    \STATE Train $f$ on this augmented data $\widehat{\mathcal{R}}_b$ and update $\theta$
    \ENDFOR
    \STATE {\bfseries return:} Final segmentation network $f$ with parameters $\theta$
\end{algorithmic}
\end{algorithm}

As shown in \cref{alg:X-Paste}, it displays the overall process of our baseline, which does not consider the different impacts each sample could impose on the model.
In other words,  our aim is to identify a function, $\phi(g,\theta)$, capable of gauging the contribution of any given generated sample $g \in \mathcal{G}$ to the current model $f$.
Then, via this scoring mechanism, we can filter and retain the most helpful samples for the model and simultaneously discard those that are useless or even harmful to the model.

\section{Our method}
\subsection{Estimation of Contribution in the Ideal Scenario}
\label{sec:ideal}
Here we first provide the ideal estimation of the contribution of an independent sample. Assuming that a target test set $\mathcal{U}$ is given, the contribution of $g \in \mathcal{G}$ to $f$ can be measured by calculating the change in the loss of $f$ on $\mathcal{U}$, that is,
\begin{equation}
    \phi(g,\theta) = L_\mathcal{U}(\theta) - L_\mathcal{U}(\theta + \Delta \theta_g)
\end{equation}
where $\Delta \theta_g$ represents the one-step gradient update on $g$, that is,
\begin{equation}
    \Delta \theta_g = -\alpha \nabla_{\theta} \ell(\mathbf{I}_g, \mathbf{Y}_g ; {\theta})
\end{equation}
where $\alpha$ represents the learning rate.

Moreover, we can employ %
the 
classic first-order Taylor expansion to approximate $\phi(g,\theta)$, which has also been widely used in previous work \cite{pruthi2020estimating,he2023sensitivity}.
Note that, it is possible to use  more sophisticated methods here, \textit{e.g.}, \citet{koh2017understanding}.
\begin{lemma}
    \label{lem:first-order}
    The loss of a network $f$ on a dataset $\mathcal{U}$ can be approximated by a first-order approximation: 
    \begin{equation}
        L_\mathcal{U}(\theta + \Delta \theta) = L_\mathcal{U}(\theta) + \Delta \theta^T \nabla_{\theta} L_\mathcal{U}(\theta) + O(\Delta \theta^2)
    \end{equation}
    \vspace*{-0.8cm}
\end{lemma}
So the contribution of $g$ to $f$ can be approximated by:
\begin{equation}
    \label{eq:approx}
    \phi(g,\theta) \approx \phi_a (g,\theta) = \alpha \nabla_{\theta} \ell(\mathbf{I}_g, \mathbf{Y}_g ; {\theta})^T \nabla_{\theta} L_\mathcal{U}(\theta)
\end{equation}

\begin{remark}
    \label{rem:1}
    \textbf{Advantages of first-order approximation:} In the case of a fixed test set $\mathcal{U}$, it eliminates the need for multiple forward computations on $\mathcal{U}$, requiring only one forward computation and one backward propagation. so it brings the possibility expand the scale of $\mathcal{U}$. 
    Besides, the fact that there is no need to update the weights makes our subsequent designs more flexible.
\end{remark}

\begin{remark}
    \label{rem:2}
    \textbf{(Test set)} In real-world scenarios, the test set $\mathcal{U}$ is typically unknown. A straightforward solution might be to reserve a part of the training set $\mathcal{R}$ as an unseen test set. However, this approach poses issues for datasets with significant category imbalance, such as LVIS, where certain long-tailed categories only appear once in the training set. %
    In light of this, we resort to a strategy that involves sampling a subset of $\mathcal{R}$ to serve as the test set $\mathcal{U}$ for each iteration. See \cref{sec:qualitative} for more details.
\end{remark}

\subsection{Batched Streaming Generative Active Learning}
\label{sec:batch}

While we have proposed an ideal estimation of contribution in the preceding section, it is not directly transferable to a real segmentation training process. The specific reasons are as follows:
\begin{enumerate}
    \item  In the previous discussion, we posit that $g \in \mathcal{G}$ is an independent sample, yet we actually paste multiple instances from $\mathcal{G}$ to $I_r$. This results in reciprocal influences among these instances, as well as interactions with $I_r$.
    \item We conduct batch data training, eliminating the possibility of estimating each instance individually, as this would provoke excessive computation.
    \item Our model undergoes constant updates. Therefore, even the same sample's contribution to the model varies under different training stages.  Furthermore, given the near-infinite data pool, the entire training process closely resembles a streaming process\cite{pmlr-v202-saran23a}. After each data entry, we must decide whether to include this data in the present update. %
    \vspace*{-0.1cm}
\end{enumerate}

For the first and second points, we redefine the contribution function $\phi(g,\theta)$ so that it can simultaneously consider the mutual influence between multiple instances and the real image $I_r$, and can also adapt to batch data training.

\begin{definition}
\label{def:inj}
Specifically, for a certain iteration $t$, we introduce a batch of data $\mathcal{R}_b$ and the corresponding instances $\mathcal{G}_b$ to be pasted. We will paste $\mathcal{G}_b$ to each image in $\mathcal{R}_b$ in a predefined random way to get $\widehat{\mathcal{R}}_b = \mathcal{R}_b \oplus \mathcal{G}_b$.
Then our contribution function $\phi(\mathcal{G}_b, \mathcal{R}_b, \theta)$ can be defined as
\begin{equation}
    \label{eq:loss_diff}
    \phi(\mathcal{G}_b, \mathcal{R}_b, \theta) = L_U(\theta + \Delta \theta_{\mathcal{R}_b}) - L_U(\theta + \Delta \theta_{\widehat{\mathcal{R}}_b} )  
\end{equation}

where $\Delta \theta_{\widehat{\mathcal{R}}_b}$ represents a one-step gradient update on $\widehat{\mathcal{R}}_b$:
\begin{equation}
    \Delta \theta_{\widehat{\mathcal{R}}_b} = - \alpha \nabla_{\theta} L_{\widehat{\mathcal{R}}_b}(\theta)
\end{equation}
Likewise, $\Delta \theta_{\mathcal{R}_b}$ represents a one-step gradient update on $\mathcal{R}_b$.
\end{definition}

In response to the third point, based on \cref{def:inj}, we can propose an algorithm called Batched Streaming Generative Active Learning (BSGAL), as shown in \cref{alg:batch}.

The basic idea is to calculate the loss and gradient of the model on $\widehat{\mathcal{R}}_b$ and $\mathcal{R}_b$ respectively and use the two updated models to calculate the loss on $\mathcal{U}_b$.  Then measure the contribution of $\mathcal{G}_b$  with the difference of two losses. Ultimately, we decide whether to accept this batch of generated data.

\begin{algorithm}[tb]
    \caption{Batched streaming generative active learning (BSGAL)}
    \label{alg:batch}
 \begin{algorithmic}[1]
     \REQUIRE labeled real data $\mathcal{R}$, generated data $\mathcal{G}$, batch size $B$, number of iterations $T$, pretrained segmentation network $f$ with parameters $\theta$, maximum number of paste instances $K$ for each image, one step learning rate $\alpha$, contribution threshold $\tau$
     \FOR{$t=1$ {\bfseries to} $T$}
     \STATE Sample a batch of real data $\mathcal{R}_b \in \mathcal{R}$ $\text{with batch size}$ $B_{accept}$.
     \FOR{$(\mathbf{I}_r , \mathbf{Y}_r) \in \mathcal{R}_b$}
     \STATE Get a random number $k$ from $[0, K]$
     \STATE Sample $k$ instances  in a class-balanced way from $\mathcal{G}$ to get $\widehat{\mathbf{I}}_r, \widehat{\mathbf{Y}}_r  = Copypaste(\mathcal{G}_b, \mathbf{I}_r)$
     \ENDFOR
     \STATE Merge all the augmented image and label pairs to get $\widehat{\mathcal{R}}_b = \mathcal{R}_b \oplus \mathcal{G}_b$
     \STATE Calculate loss on $\mathcal{R}_b$ and $\widehat{\mathcal{R}}_b$  and get the gradient $ \nabla_{\theta} L_{\mathcal{R}_b}(\theta)$ and $\nabla_{\theta} L_{\widehat{\mathcal{R}}_b}(\theta)$
     \STATE Sample a batch of data $\mathcal{U}_b \in \mathcal{R}$ as test set with batch size $B_{test}$
     \STATE Calculate the contribution $\phi(\mathcal{G}_b, \mathcal{R}_b, \theta)$ using \cref{eq:loss_diff}
     \IF {$\phi(\mathcal{G}_b, \mathcal{R}_b, \theta) > \tau$}
        \STATE Train $f$ on this augmented data $\widehat{\mathcal{R}}_b$ and update $\theta$
     \ELSE 
        \STATE Train $f$ on this real data $\mathcal{R}_b$ and update $\theta$
     \ENDIF
     \ENDFOR
     \STATE {\bfseries return:} Final segmentation network $f$ with parameters $\theta$
 \end{algorithmic}
 \end{algorithm}

As pointed out in \cref{rem:1}, there are some advantages of using one-order approximation.
Building upon \cref{lem:first-order}, we can further approximate $\phi(\mathcal{G}_b, \mathcal{R}_b, \theta)$ as
\begin{equation}
    \vspace*{-0.1cm}
    \label{eq:grad}
    \phi(\mathcal{G}_b, \mathcal{R}_b, \theta) \approx \alpha \nabla_{\theta} ( L_{\widehat{\mathcal{R}}_b}(\theta) - L_{\mathcal{R}_b}(\theta) )^T \nabla_{\theta} L_U(\theta)
    \vspace*{-0.1cm}
\end{equation}

So the \cref{alg:batch} can be further simplified by using \cref{eq:grad} in Line 10.

\begin{remark}
    \label{rem:batch}
    \textbf{(Batch size)}
    It is crucial to note that three distinct batch sizes here, one is the batch size $B_{train}$ for model update, one is the batch size $B_{accept}$ for calculating the data contribution to determine whether to accept, and the other is the batch size $B_{test}$ of the test set formed by sampling.  These three are not necessarily identical.
    Ideally, a smaller $B_{accept}$ is preferable – when the batch size is reduced to 1, it allows for a more accurate per-image estimation.
    In the actual implementation process, we execute this algorithm independently on each GPU, except for summing up all losses when updating the model. 
    Consequently, the batch size $B_{train} = B_{accept} \times \#GPU$. 
\end{remark} 

Usually, due to the limitation of GPU memory, $\mathcal{U}_b$ can only take a very small batch size $B_{test}$, which will lead to the instability and inaccuracy of the estimation of $\phi(\mathcal{G}_b, \mathcal{R}_b, \theta)$.
Thanks to the one-order approximation, we can consider more test data when calculating $\phi(\mathcal{G}_b, \mathcal{R}_b, \theta)$.
Specifically, we can keep a grad cache to record the grad obtained on other batches in the previous iterations. Every time we get the current batch $\nabla_{\theta} L_{\mathcal{U}_b}(\theta)$, we will update the grad cache in a momentum way.
Then when calculating $\phi(\mathcal{G}_b, \mathcal{R}_b, \theta)$, we use the grad in the grad cache to replace the grad of the current batch, which is equivalent to expanding the scale of $\mathcal{U}_b$, thereby achieving a more stable estimation. 

However, this batch size $B_{test}$ is not the bigger the better. Our subsequent experiments prove that if we use the grad cache to approximate the entire training set $\mathcal{R}$ as $\mathcal{U}_b$, it will lead to a decrease in the performance of the model. We believe that this is caused by overfitting. When we approximate the entire training set $\mathcal{R}$ as $\mathcal{U}_b$, we can only screen out samples similar to the training set $\mathcal{R}$,  thereby inhibiting the diversity of the data. 
Diversity is also an important factor considered in many active learning works. 
We believe that the design here parallels the idea of diversity-based active learning \cite{sener2018active,geifman2017deep} using a core-set, with the only difference being that core-set methods mine the most representative samples from unlabeled data, while we randomly sample some from the labeled real data to act as a kind of ``core-set''.
In our task, the diversity of generated data is particularly crucial for long-tailed categories, which can effectively bridge the gap between scarce training data and real-world distribution. Therefore, we need to balance the stability of the estimation and the diversity of the data. Thus, we use the momentum method to update the grad cache, which can ensure the stability of the estimation to a certain extent, and at the same time will not inhibit the diversity of the data.

The modified contribution estimation algorithm for the final BSGAL is shown in \cref{alg:batch3}.
\begin{algorithm}[tb]
    \caption{ Contribution estimation based on gradient cache  }
    \label{alg:batch3}
 \begin{algorithmic}[1]
     \REQUIRE gradient $\nabla_{\theta} L_{\mathcal{R}_b}(\theta)$,  gradient $ \nabla_{\theta} L_{\widehat{\mathcal{R}}_b}(\theta)$, momentum coefficient $\beta$, current iteration $t$, model parameters $\theta$, sampled test set $\mathcal{U}_b$
     \STATE Calculate loss on $\mathcal{U}_b$ to get $L_{\mathcal{U}_b}$ and get the grad $\nabla_{\theta} L_{\mathcal{U}_b}(\theta)$
     \IF {$t == 1$}
        \STATE Initialize grad cache $\mathbf{C} = \nabla_{\theta} L_{\mathcal{U}_b}(\theta)$
     \ELSE 
        \STATE Update the grad cache with the grad of current batch $\mathbf{C} = \beta \mathbf{C} + (1 - \beta) \nabla_{\theta} L_{\mathcal{U}_b}(\theta)$
     \ENDIF
     \STATE Calculate the contribution of $\mathcal{G}_b$ as $\phi(\mathcal{G}_b, \mathcal{R}_b, \theta) = \alpha \nabla_{\theta} ( L_{\widehat{\mathcal{R}}_b}(\theta) - L_{\mathcal{R}_b}(\theta) )^T \mathbf{C}$
     \STATE {\bfseries return:} $\phi(\mathcal{G}_b, \mathcal{R}_b, \theta)$
 \end{algorithmic}
 \end{algorithm}

 \begin{remark}
    \label{rem:off}
    \vspace*{-0.1cm}
    \textbf{(Extension to offline learning)}
    Offline learning needs to satisfy the following two assumptions: 1. The generated data is limited or small in scale. 2. The model parameters will not change significantly during the fine-tuning process.
    Our method can also be easily extended to the case of offline learning. Specifically, we only need to use a fixed model, and then use the entire dataset $\mathcal{R}$ as the test set $\mathcal{U}$. Forward and backward once on the entire dataset, we can get the gradient $\nabla_{\theta} L_{\mathcal{U}}(\theta)$ for the entire dataset. Then we can use this gradient to calculate the contribution of each generated sample.
    Using \cref{eq:approx}, we can estimate the contribution of each generated sample.
\end{remark}

\section{Experiments}
\label{sec:exp}
First, we perform some analytical experiments in an offline setting(as discussed in \cref{rem:off}) to verify the feasibility of our method and also to facilitate a better understanding of our method for readers. 
Then, we conduct the main experiments under the online setting, compared with our baseline. Key ablation studies are also conducted to substantiate the efficiency of our method. Detailed information about the implementation can be found in \cref{app:implement}.
\subsection{Offline Setting}

\subsubsection{ CIFAR-10}
\label{sec:toy}
In this section, we conduct a toy experiment on  CIFAR-10 \cite{krizhevsky2009learning} to verify our method. The original CIFAR-10 dataset comprises five splits within the training set, each containing 10,000 images, and a test set equally housing 10,000 images.
We use the first split in the training set as our training set $\mathcal{R}$, and the remaining 4 splits are added with noise of different scales (40,100,200,400) to simulate the generated data $\mathcal{G}$. 
We use the model trained only on the first split to perform offline mining and then use 1000 images in the first split as test set  $\mathcal{U}$. By estimating the contribution of each sample, we can draw the distribution of the contribution of samples on different splits.\textbf{}

\begin{figure}
    \centering
    \includegraphics[width=0.98\linewidth]{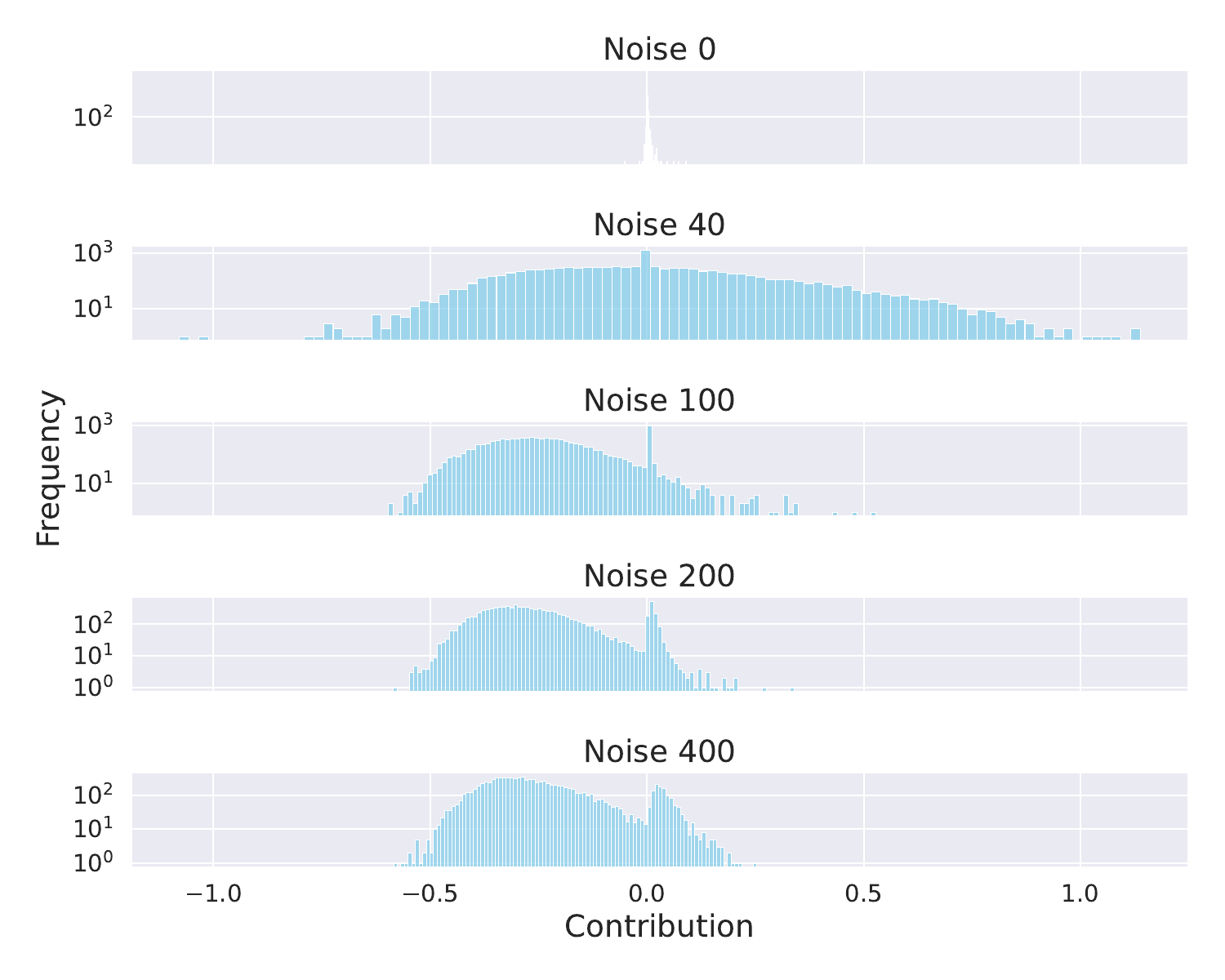}
    \caption{%
    The distribution of contributions under different noise scales.
    }
    \label{fig:grad_sim_cifar}
\end{figure}

As shown in \cref{fig:grad_sim_cifar},  it is observable that with the escalating scale of noise, the distribution of contributions progressively shifts to the left.  This indicates that excessive noise tends to negatively impact the model. Note that the split with a noise of 0 is our training set, so we can see that the contribution values of these samples are concentrated around zero. 
In other words, these samples can no longer bring positive effects to the model because they have been fully utilized in previous training. %
This observation is consistent with some previous active learning work \cite{cai2013maximizing,ash2021gone,pmlr-v202-saran23a}, where they also estimate the amount of information or the difficulty level of samples through gradients. However, they do not consider the positive or negative contributions but only select samples with larger absolute values. We further conduct quantitative experiments, as shown in \cref{tab:cifar10}, to prove that using our method to select data can effectively improve the performance of the model.

\begin{table}[t]
\vspace*{-0.3cm}
\centering
\caption{Using our method to select samples brings improvement to the model.}
\vspace*{0.3cm}
\label{tab:cifar10}
\begin{tabular}{@{}lc@{}}
\toprule
Training set                                        & Accuracy (avg $\pm$ std) \\ \midrule
only $\mathcal{R}$                      & 86.28 $\pm$ 0.55         \\
$\mathcal{R}$ + all $\mathcal{G}$       & 86.71 $\pm$ 0.21         \\
$\mathcal{R}$ +  selected $\mathcal{G}$ & 87.61 $\pm$ 0.20         \\ \bottomrule
\end{tabular}
\end{table}

\subsubsection{LVIS}
\label{sec:qualitative}
\begin{figure}[h]
    \centering
    \includegraphics[width=0.995\linewidth]{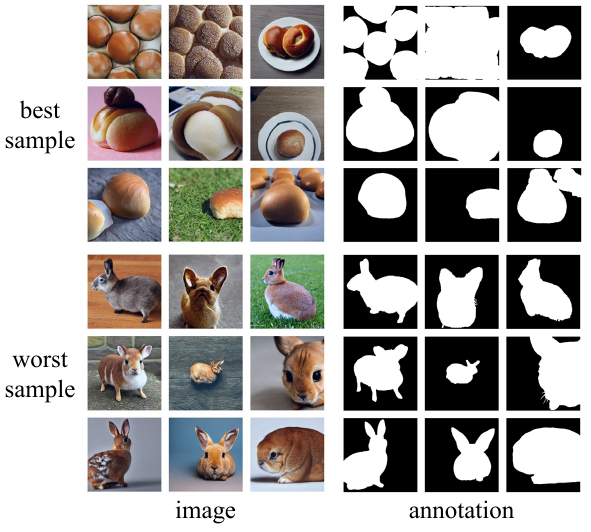}
    \caption{The best and worst samples found using our contribution estimation function for a LVIS class `bun'.}
    \label{fig:grad_sim_lvis}
    \vspace*{-0.2cm}
\end{figure}

\begin{figure}[h]
    \centering
    \includegraphics[width=0.9\linewidth]{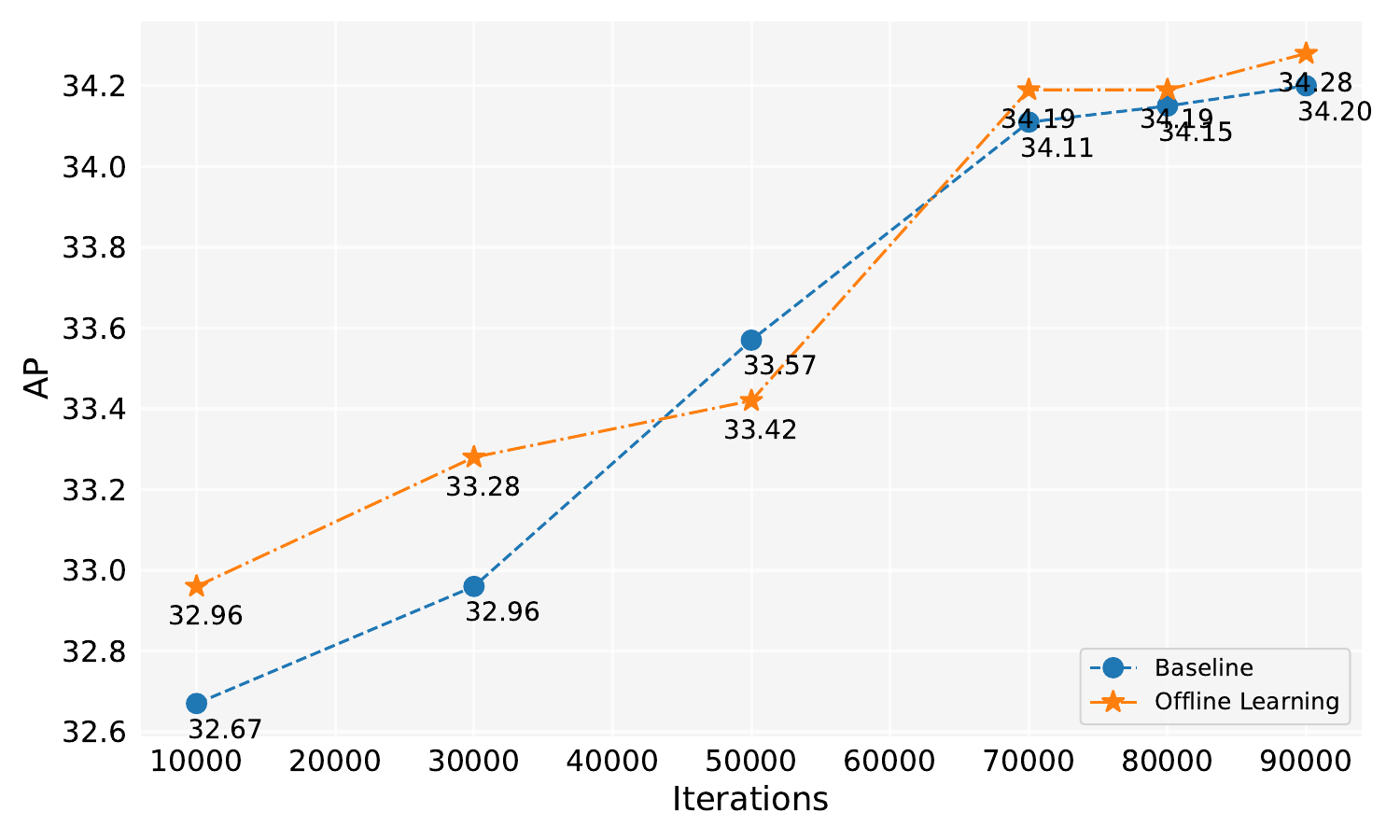}
    \caption{%
    Performance of the model under different iterations.}
    \vspace*{-0.2cm}
    \label{fig:offline_iter}
\end{figure}
We further carry out offline experiments on the generated data of LVIS categories, qualitatively examining the efficacy of our method. 
We calculate the gradient $\nabla_{\theta} L_{\mathcal{U}}(\theta)$  pre-emptively on the LVIS training set with a trained model.

\begin{table*}[t]
\centering
\vspace*{-0.3cm}
\caption{Main results on LVIS. ``+CLIP'' means using CLIP to filter the generated data.}
\vspace*{0.1cm}
\label{tab:main}
\begin{tabular}{@{}lccccccccc@{}}
\toprule
Method          & Backbone & $AP^{box}$     & $AP^{mask}$    & $AP_r^{box}$   & $AP_r^{mask}$  & $AP_c^{box}$   & $AP_c^{mask}$  & $AP_f^{box}$   & $AP_f^{mask}$  \\ \midrule
Baseline        & R50      & 34.20          & 30.39          & 24.33          & 22.21          & 33.23          & 29.57          & 39.63          & 34.89          \\
Baseline + CLIP & R50      & 34.35          & 30.70          & 25.99          & 24.38          & 32.83          & 29.41          & 39.71          & 34.92          \\
Ours            & R50      & \textbf{35.40} & \textbf{31.56} & \textbf{27.95} & \textbf{25.43} & \textbf{34.14} & \textbf{30.56} & \textbf{40.07} & \textbf{35.37} \\
\midrule
Baseline        & Swin-L   & 49.57          & 43.85          & 44.87          & 39.66          & 49.74          & 44.64          & 51.46          & 44.82          \\
Baseline + CLIP & Swin-L   & 49.80          & 44.51          & 45.28          & 40.62          & 49.33          & 44.96          & \textbf{52.30} & \textbf{45.72} \\
Ours            & Swin-L   & \textbf{50.47} & \textbf{44.85} & \textbf{47.55} & \textbf{42.37} & \textbf{50.43} & \textbf{45.47} & 51.79          & 45.26          \\ \bottomrule
\vspace*{-0.8cm}
\end{tabular}%
\end{table*}

This gradient then serves to estimate each instance's contribution. Subsequently, we rank these instances in decreasing order of their contribution, facilitating per-image analysis. As an illustrative example, we use a `bun' category from the LVIS, because we discover that Stable Diffusion does not perform optimally within this category, often leading to confusion between `bun' and `bunny', thereby resulting in the generation of ambiguous data. As depicted in \cref{fig:grad_sim_lvis}, it can be observed that the instances having the most significant contributions are nearly unambiguous, whereas the instances with minimal contributions are mostly incorrect, resulting in rabbit images being generated.
Therefore, through our method, we can effectively filter out the generated data with ambiguity.

To verify the indispensability of online learning, we first use the offline method to filter the generated data for training and compare it with our baseline. As shown in \cref{fig:offline_iter}, the offline method can only bring a slight improvement to the final model performance.
In addition, in the early stage of model training, this performance improvement is still quite obvious, but with the training process, this performance improvement gradually diminishes.
We conjecture that this trend is likely due to the offline contribution estimation's reliance on the initial model, and as the model undergoes training, the parameters change significantly, which leads to the inaccuracy of the offline contribution estimation. Therefore,  the necessity arises for online contribution estimation.

\subsection{Online Setting}
\subsubsection{Main results}
\label{sec:main}

To validate the effectiveness of our method in handling long-tailed segmentation tasks, we perform experiments on the LVIS \cite{gupta2019lvis} dataset.  A strong baseline —— X-Paste \cite{zhao2022XPaste}, is compared with our method. We further examine the impact of the usage (or non-usage) of CLIP \cite{radford2021learning}, as mentioned in their paper, for filtering generated data. Given that X-Paste does not open source the generated data used, we have re-implemented the data generation pipeline and generated thousands of images for each category of LVIS. The specific generation details are shown in the appendix. In addition, original X-Paste also uses retrieval to obtain extra real data and uses Copy-Paste to augment the real training data. Since the primary focus of our methodology is on the selection of generated data, to mitigate the influences from those additional factors, we choose to refrain from using these tricks.

As shown in \cref{tab:main}, following X-Paste, the segmentation architecture we used is CenterNet2~\cite{zhou2021probabilistic}, and we test two different backbones, ResNet50 \cite{he2016deep} and Swin-L \cite{liu2021swinv2}.
It can be observed that the impact of filtration via CLIP is rather subtle, with the $AP^{box}$ witnessing an increment of merely 0.1 $\sim$ 0.2 points. 
Furthermore, 
we also find that CLIP filtering %
shows 
a more significant improvement on $AP^{mask}$. 
Conversely, our method continues to deliver substantial improvements across all categories when contrasted with the use of CLIP filtration. 
This improvement is especially notable  for 
long-tailed categories, $AP_r^{box}$ increases by 2 $\sim$ 2.3, and $AP_r^{mask}$ increases by 1 $\sim$ 1.7.

\subsubsection{Ablation}
\label{sec:qualitative}
\begin{table}[h]
\vspace*{-0.3cm}
\centering
\caption{Comparison of different $L_{test}$ on contribution estimation.}
\label{tab:loss}
\vspace*{0.3cm}
\begin{tabular}{@{}ccccc@{}}
\toprule
Loss       & $AP^{box}$     & $AP^{mask}$    & $AP_r^{box}$   & $AP_r^{mask}$  \\ \midrule
cls       & \textbf{35.24} & \textbf{31.49} & \textbf{28.14} & \textbf{25.74} \\
cls stage0 & 34.94          & 31.23          & 26.34          & 24.27          \\
all        & 34.98          & 30.94          & 26.91          & 23.87          \\ \bottomrule
\end{tabular}%

\end{table}

\begin{table*}[t]
\vspace*{-0.5cm}
\centering
\caption{Comparison of different algorithms we designed.}
\label{tab:estimate}
\vspace*{0.1cm}
\begin{tabular}{@{}ccccccccc@{}}
\toprule
Method                                             & $AP^{box}$     & $AP^{mask}$    & $AP_r^{box}$   & $AP_r^{mask}$  & $AP_c^{box}$   & $AP_c^{mask}$  & $AP_f^{box}$   & $AP_f^{mask}$  \\ \midrule
Loss estimate (\cref{alg:batch})  & 35.11          & 31.29          & 27.47          & 25.10          & 33.55          & 30.14          & \textbf{40.20} & 35.30          \\
Grad estimate (\cref{eq:grad}) & 35.24          & 31.49          & \textbf{28.14} & \textbf{25.74} & 33.67          & 30.31          & 40.12          & 35.34          \\
Grad cache (\cref{alg:batch3})    & \textbf{35.40} & \textbf{31.56} & 27.95          & 25.43          & \textbf{34.14} & \textbf{30.56} & 40.07          & \textbf{35.37} \\
Grad cache (global)                                & 35.07          & 31.38          & 26.84          & 24.82          & 33.74          & 30.36          & 40.18          & \textbf{35.40} \\ \bottomrule
\end{tabular}%
\vspace*{-0.5cm}
\end{table*}
\textbf{Loss.}   
The algorithm in \cref{sec:batch} deploys two distinct types of loss. First is the loss used to train and finally update the model, $L_{train}$, which is subject to the original network design. In our case, We utilize the loss in CenterNet2 \cite{zhou2021probabilistic}.
For the instance segmentation task, the actual $L_{train}$ is composed of multiple tasks and multiple stages of Loss. Specifically, it can be written in the following form\footnote{
    In CenterNet2, there is only one stage for mask loss. In addition, there are some other losses, which are not listed for simplicity.
}:
\begin {equation}
    \label{eq:loss}
    \vspace*{-0.3cm}
    L_{train} = \sum_{i=1}^{S} (L_{cls}^{i} + L_{box}^{i} + L_{mask}^{i})
    \vspace*{-0cm}
\end {equation}
where $S$ represents the number of stages, $L_{cls}^{i}$, $L_{box}^{i}$, $L_{mask}^{i}$ represent the classification loss, regression loss, and segmentation loss of the $i$-th stage, respectively.

Second is the test loss $L_{test}$ used to estimate the contribution, which does not need to be consistent with $L_{train}$.
Considering that the main purpose of introducing generated data is to improve classification on long-tailed categories, $L_{cls}$ may play a dominant role. Thus, we compare the effects of different $L_{test}$ on contribution estimation in \cref{tab:loss}.

In this experiment, we are based on \cref{eq:grad}, where ``cls'' denotes using all stages of $L_{cls}$  ``cls stage0'' denotes only using the first stage of $L_{cls}$, and ``all'' denotes using all Loss in $L_{train}$.
The findings indicate that if only the $L_{cls}$ of first stage is used, the efficiency of contribution estimation diminishes. Additionally, utilizing all losses within $L_{train}$ does not surpass the performance of solely exploiting $L_{cls}$, which also verifies our assumption that $L_{cls}$ dominates in contribution estimation, and other losses may cause interference. Therefore, our final decision is to rely singularly on $L_{cls}$ for estimating contribution.

\textbf{Contribution estimation.}
 We are interested in whether the three algorithms proposed in \cref{sec:batch} are effective. Therefore, we conduct comparative experiments here, and the specific results are presented in \cref{tab:estimate}.

We observe that there's not a significant difference in performance among the three algorithms, all of which demonstrated efficacy when compared to the baseline.  Overall, \cref{alg:batch3} has the best effect, which is also the algorithm we finally adopted. 
Hence, our proposed method of using a larger test set $\mathcal{U}_b$, ensuring a smoother and more stable contribution estimation by updating the grad cache with momentum, is proven effective.
Compared with Grad estimate and Loss estimate, it can be proved that the first-order approximation will not bring significant performance loss. 

As discussed in \cref{sec:batch}, there exists a trade-off between the stability of the estimation and the diversity of the filtered data. To verify this, we additionally added the fourth experiment, where we use global average pooling to estimate the contribution. Specifically,
We modify the original update method $\mathbf{C} = \beta \mathbf{C} + (1 - \beta) \nabla_{\theta} L_{\mathcal{U}_b}(\theta)$ 
to be related to the current iteration $t$, $\mathbf{C} = \frac{t-1}{t} \mathbf{C} + \frac{1}{t} \nabla_{\theta} L_{\mathcal{U}_b}(\theta)$.
That is, $\mathbf{C} = \frac{1}{t} \sum_{i=1}^{t} \nabla_{\theta_i} L_{\mathcal{U}_b^i}(\theta_i)$.

However, this modification incurs a slight performance drop. We believe that this is due to enforcing the generated data to align with the distribution of the entire training set and suppressing the diversity of the data.
For long-tailed categories with relatively few real data, this diversity is quite significant. That's why using global average pooling, $AP_r^{box}$ and $AP_r^{mask}$ have a significant drop, while $AP_f^{mask}$ exhibits an improvement.
Correspondingly, Grad estimate, which solely uses the current batch to estimate the contribution of the test set,  is most conducive to ensuring data diversity, so the performance of $AP_r^{box}$ and $AP_r^{mask}$ is the best.

\begin{table}[h]
    \centering
    \caption{Comparison of different sampling strategies for $\mathcal{U}_b$.}
    \vspace*{0.1cm}
    \label{tab:sampling}
    \begin{tabular}{@{}lcccc@{}}
\toprule
Strategy & $AP^{box}$     & $AP^{mask}$    & $AP_r^{box}$   & $AP_r^{mask}$  \\ \midrule
all classes       & 35.21          & 31.38          & 26.75          & 24.21          \\
pasted classes    & \textbf{35.40} & \textbf{31.56} & \textbf{27.95} & \textbf{25.43} \\
all images        & 35.15          & 31.26          & 26.59          & 23.75          \\ \bottomrule
\end{tabular}%
\end{table}

\textbf{Sampling strategy.} We compare three different sampling test set strategies:
1. Sample from all classes: Sample uniformly from all categories, and then sample uniformly from the image pool corresponding to the sampled category.
2. Sample from pasted classes: Sample uniformly from the categories in the generated data $\mathcal{G}_b$ used in this batch, and then sample uniformly from the image pool corresponding to the sampled category.
3. Sample from all images: Sample uniformly from all image pools. 

\begin{table}[h]
    \vspace*{-0.3cm}
    \centering
    \caption{Comparison of random batch-level dropout and our method.}
    \vspace*{0.2cm}
    \label{tab:dropout}
    \setlength{\tabcolsep}{3pt}
    \begin{tabular}{@{}lcccc@{}}
    \toprule
    Method              & $AP^{box}$     & $AP^{mask}$    & $AP_r^{box}$   & $AP_r^{mask}$  \\ \midrule
    Random Dropout      & 34.73          & 30.96          & 25.05          & 22.69          \\
    Our method          & \textbf{35.40} & \textbf{31.56} & \textbf{27.95} & \textbf{25.43} \\ \bottomrule
    \end{tabular}%
    \vspace*{-0.1cm}
\end{table}
As indicated in \cref{tab:sampling}, it is evident that the approach of uniformly sampling from pasted categories delivers the most effective performance, thus we finalize on this sampling strategy. Especially in $AP_r^{box}$ and $AP_r^{mask}$, the improvement of this sampling strategy is the most obvious. Compared with sampling from all images, sampling from class can better ensure the balance of categories, thereby enhancing the impact on rare categories.
As for the comparison with sampling from all classes, sampling from pasted classes is more directional, so the estimation of the contribution is more accurate, leading to a boost in overall performance.

\textbf{Random batch-level dropout.}
Our algorithm is essentially to accept or reject the generated data on a batch-by-batch basis. Therefore, when the contribution evaluation of the data is completely invalid, our algorithm degenerates into a random batch-level dropout.
To verify that it is not this random dropout that brings performance improvement, we conduct a random batch-level Dropout experiment with the same acceptance rate. %
\cref{tab:dropout} shows that although random dropout will also bring a slight improvement, compared with our method, there is still a very obvious gap, which shows that the improvement brought by our method is not entirely due to random dropout.

\section{Conclusion}
In this paper, we propose a new problem, how to design an effective method to realize the effective screening and utilization of generated data, to further improve the performance of downstream perception tasks.
To address this problem, we propose a gradient-based generated data contribution estimation method and embed it into the actual training process. We design a complete pipeline that can automatically generate data to improve the performance of downstream perception tasks. Experiments prove that our method can achieve better performance than unfiltered or CLIP-filtered methods on long-tailed segmentation tasks.
\section*{Broader Impact}
Our 
goal is to advance the field of Machine Learning. There are many potential societal consequences of our work, none which we feel must be specifically highlighted here.

\bibliography{example_paper}
\bibliographystyle{ieeenat_fullname}

\newpage
\appendix
\onecolumn
\section{Implementation Details}
\label{app:implement}
\subsection{Dataset}
We choose LVIS~\cite{gupta2019lvis} as the dataset for our experiments. LVIS is a large-scale instance segmentation dataset, comprising approximately 160,000 images with over 2 million high-quality instance segmentation annotations across 1203 real-world categories. The dataset is further divided into three categories: rare, common, and frequent, based on their occurrence across images. Instances marked as `rare' appear in 1-10 images, `common' instances appear in 11-100 images, whereas `frequent' instances appear in more than 100 images. The overall dataset exhibits a long-tail distribution, closely resembling the data distribution in the real world, and is widely applied under multiple settings, including few-shot segmentation 
\cite{liu2023matcher} and open-world segmentation \cite{wang2022open,zhu2023segprompt}. Therefore, we believe that selecting LVIS allows for a better reflection of the model's performance in real-world scenarios. We use the official LVIS dataset splits, with about 100,000 images in the training set and 20,000 images in the validation set.

\subsection{Data Generation}
Our data generation and annotation process is consistent with \citet{zhao2022XPaste}, and we briefly introduce it here.
We first use StableDiffusion V1.5~\cite{Rombach_2022_CVPR} (SD) as the generative model. For the 1203 categories in LVIS~\cite{gupta2019lvis}, we generate 1000 images per category, with image resolution 512~$\times$~512. The prompt template for generation is ``a photo of a single \textit{\{CATEGORY\_NAME\}}''. We use U2Net~\cite{qin2020u2}, SelfReformer~\cite{yun2022selfreformer}, UFO~\cite{su2023unified}, and CLIPseg~\cite{luddecke2022image} respectively to annotate the raw generative images, and select the mask with the highest CLIP score as the final annotation. To ensure data quality, images with CLIP scores below 0.21 are filtered out as low-quality images. During training, we also employ the instance paste strategy provided by \citet{zhao2022XPaste} for data augmentation. For each instance, we randomly resize it to match the distribution of its category in the training set. The maximum number of pasted instances per image is set to 20. 

In addition, to further expand the diversity of generated data and make our research more universal, we also use other generative models, including DeepFloyd-IF~\cite{alex2023deepfloyd} (IF) and Perfusion \cite{Tewel2023KeyLockedRO} (PER), with 500 images per category per model.
For IF, we use the pre-trained model provided by the author, and the generated images are the output of Stage II, with a resolution of 256$\times$256.
For PER, the base model we use is StableDiffusion V1.5. For each category, we fine-tune the model using the images croped from the training set, with 400 fine-tuning steps. We use the fine-tuned model to generate images.

\begin{table}[h]
\centering
\caption{Comparison of different generated data.}
\label{tab:genmodel}
\begin{tabular}{@{}cccccccccc@{}}
\toprule
Method   & Generated Data & $AP^{box}$     & $AP^{mask}$    & $AP_r^{box}$   & $AP_r^{mask}$  & $AP_c^{box}$   & $AP_c^{mask}$  & $AP_f^{box}$   & $AP_f^{mask}$  \\ \midrule
Baseline & SD             & 34.00          & 30.33          & 24.48          & 22.65          & 32.71          & 29.27          & 39.62          & 34.89          \\
Baseline & SD + IF        & 34.15          & 30.39          & 26.12          & 23.76          & 32.40          & 28.97          & 39.62          & 34.89          \\
Baseline & SD + IF + Per  & 34.20          & 30.39          & 24.33          & 22.21          & 33.23          & 29.57          & 39.63          & 34.89          \\
BSGAL    & SD             & 34.82          & 31.21          & 26.76          & 24.84          & 33.28          & 30.01          & 40.08          & 35.34          \\
BSGAL    & SD + IF        & 35.13          & 31.34          & 26.83          & 24.32          & 33.92          & \textbf{30.57} & \textbf{40.13} & 35.29          \\
BSGAL    & SD + IF + Per  & \textbf{35.40} & \textbf{31.56} & \textbf{27.95} & \textbf{25.43} & \textbf{34.14} & 30.56          & 40.07          & \textbf{35.37} \\ \bottomrule
\end{tabular}
\end{table}
We also explore the effect of using different generated data on the model performance (see \cref{tab:genmodel}). We can see that based on the original StableDiffusion V1.5, using other generative models can bring some performance improvement, but this improvement is not obvious.
Specifically, for specific frequency categories, we found that IF has a more significant improvement for rare categories, while PER has a more significant improvement for common categories. This is likely because IF data is more diverse, while PER data is more consistent with the distribution of the training set.
Considering that the overall performance has been improved to a certain extent, we finally adopt the generated data of SD + IF + PER for subsequent experiments.
\subsection{Model Training}

Follow \citet{zhao2022XPaste},  We use CenterNet2~\cite{zhou2021probabilistic} as  our segmentation model, with ResNet-50~\cite{he2016deep} or Swin-L~\cite{liu2021swinv2}  as the backbone.
For ResNet-50, the maximum training iteration is set to 90,000 and the model is initialized with weights first pretrained on ImageNet-22k then finetuned on LVIS~\cite{gupta2019lvis}, as \citet{zhao2022XPaste} did. And we use 4 Nvidia 4090 GPUs with a batch size of 16 during training.
As for Swin-L, the maximum training iteration is set to 180,000 and the model is initialized with weights pretrained on ImageNet-22k, since our early experiments show that this initialization can bring a slight improvement compared to the weights trained with LVIS.
And we use 4 Nvidia A100 GPUs with a batch size of 16 for training.  Besides, due to the large number of parameters of Swin-L, the additional memory occupied by saving the gradient is large, so we actually use the algorithm in \cref{alg:batch}.

The other unspecified parameters also follow the same settings as X-Paste~\cite{zhao2022XPaste}, such as the AdamW~\cite{loshchilov2017decoupled} optimizer with an initial learning rate of 
1e$-4$.

\subsection{Data Amount} 

\begin{figure}
    \centering
    \includegraphics[width=1\linewidth]{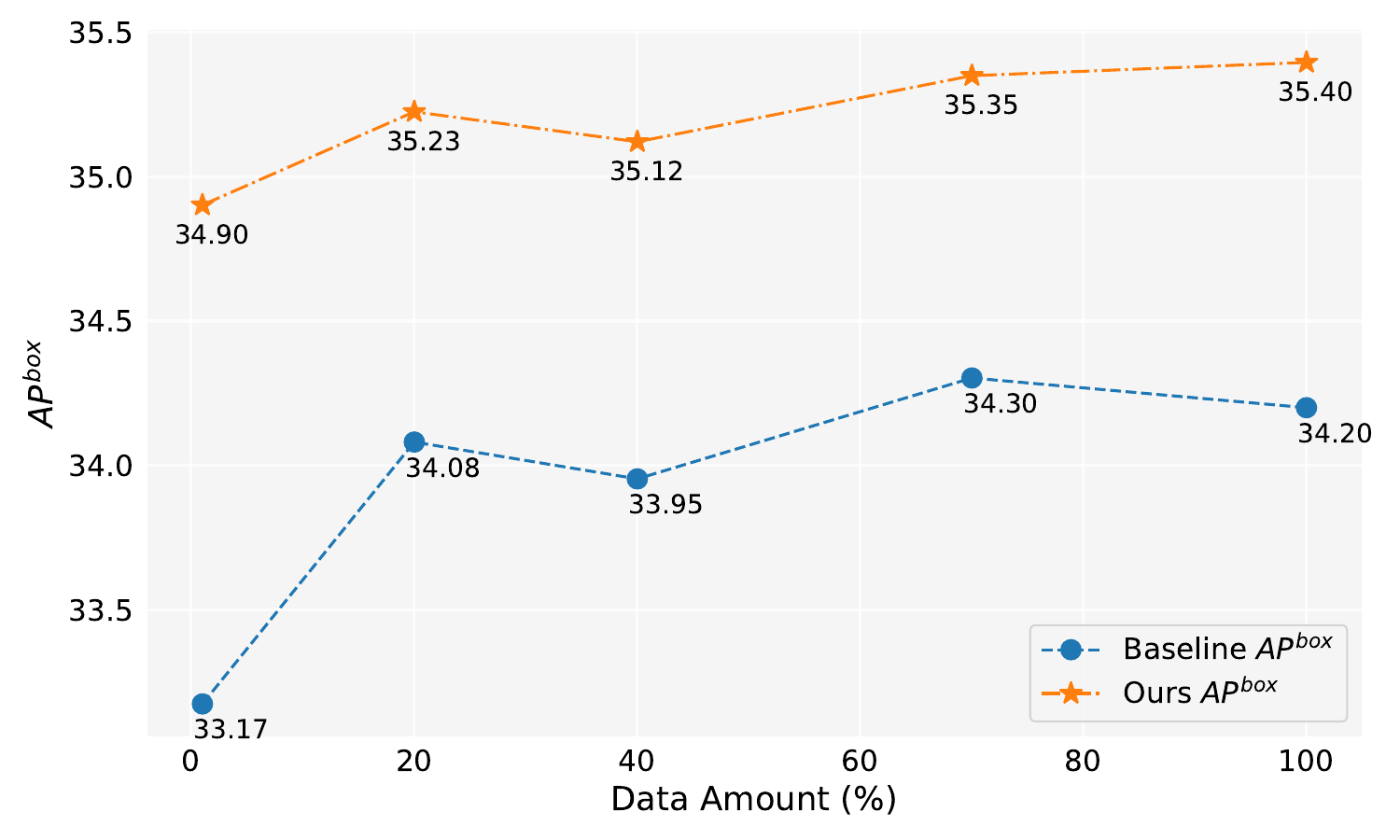}
    \caption{Model performances when using different amount of generated data.}
    \label{fig:ap_dataamount}
\end{figure}

In this work, we have generated over 2 million images. \cref{fig:ap_dataamount} shows the model performances when using different amount of generated data(1\%,10\%,40\%,70\%,100\%). 
Overall, as the amount of generated data increases, the performance of the model also improves, but there is also some fluctuation.
Our method is always better than the baseline, which proves the effectiveness and robustness of our method.

\subsection{Contribution Estimation}
As mentioned in \cref{sec:batch}, we use $ \nabla_{\theta} ( L_{\widehat{\mathcal{R}}_b}(\theta) - L_{\mathcal{R}_b}(\theta) )^T \nabla_{\theta} L_{\mathcal{U}_b}(\theta) $ to estimate the contribution of $\mathcal{G}_b$.

Here we actually consider both the direction and the magnitude of the gradient.
It is worth mentioning that many previous works actually mainly consider the magnitude of the gradient, for example, the data with large gradient magnitude has more information and should be annotated \cite{cai2013maximizing} or the wrong outlier data should be filtered \cite{pruthi2020estimating}.

If we only consider the direction, it is equivalent to normalizing each gradient first, and then calculating, 
then our calculation formula becomes $\mathcal{G}_b = \frac{\nabla_{\theta} ( L_{\widehat{\mathcal{R}}_b}(\theta) - L_{\mathcal{R}_b}(\theta) )^T \nabla_{\theta} L_{\mathcal{U}_b}(\theta)}{\|\nabla_{\theta} ( L_{\widehat{\mathcal{R}}_b}(\theta) - L_{\mathcal{R}_b}(\theta) )\|_2 \|\nabla_{\theta} L_{\mathcal{U}_b}(\theta)\|_2}$.

Thus, we essentially calculate the cosine similarity. Then we conducted an experimental comparison, as shown in \cref{tab:cosine}, we can see that if we normalize the gradient, our method will have a certain improvement.
In addition, since we need to keep two different thresholds, it is difficult to ensure the consistency of the acceptance rate. So we adopt a dynamic threshold strategy, pre-set an acceptance rate, maintain a queue to save the contribution of the previous iter, and then dynamically adjust the threshold according to the queue, so that the acceptance rate stays at the pre-set acceptance rate.

\begin{table}[]
    \centering
    \caption{Comparison of using grad normalization or not.}
    \label{tab:cosine}
    \begin{tabular}{@{}ccccccccc@{}}
\toprule
Normalize & $AP^{box}$     & $AP^{mask}$    & $AP_r^{box}$   & $AP_r^{mask}$  & $AP_c^{box}$   & $AP_c^{mask}$  & $AP_f^{box}$   & $AP_f^{mask}$  \\ \midrule
$\times$  & 35.05          & 31.27          & 26.78          & 24.34          & 33.79          & 30.32          & 40.10          & 35.39          \\
$\surd$   & \textbf{35.40} & \textbf{31.56} & \textbf{27.95} & \textbf{25.43} & \textbf{34.14} & \textbf{30.56} & \textbf{40.07} & \textbf{35.37} \\ \bottomrule
\end{tabular}
    \end{table}

\subsection{Toy Experiment}
\begin{figure*}\centering\includegraphics[width=0.9\linewidth]{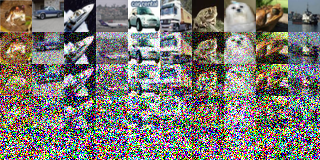}\caption{Illustration of noisy images exhibiting various noise scales and categories. Each row, from top to bottom, signifies different noise levels, specifically 0, 40, 100, 200, and 400, respectively. All images are sourced from the CIFAR-10 dataset.}\label{fig:noised_imgs}\end{figure*}

The following are the specific experimental settings implemented on CIFAR-10: We employed a simple ResNet18 as the baseline model and conducted training over 200 epochs, and the accuracy after training on the original training set is 93.02\%. The learning rate is set at 0.1, utilizing the SGD optimizer. A momentum of 0.9 is in effect, with a weight decay of 5e-4. We use a cosine annealing learning rate scheduler.
The constructed noisy images are depicted in \cref{fig:noised_imgs}. A decline in image quality is observed as the noise level escalates. Notably, when the noise level reaches 200, the images become significantly challenging to identify. For \cref{tab:cifar10}, we use Split1 as $R$, while $G$ consists of `Split2 + Noise40', `Split3 + Noise100', `Split4 + Noise200', 

\subsection{A Simplification Only  Forward Once}
In \cref{sec:batch}, we actually need one more forward on $\mathcal{R}_b$ compared to our baseline.
However, we can simplify it to only one forward. The specific reason is that as mentioned in \cref{tab:loss}, we only use the classification loss, $L_{cls}$ 
this loss is actually the sum of the cross entropy loss of each instance, and whether this instance is generated or real is known during the training process.
so the loss can be further distangled as $L_{cls} = L_{real} + L_{gen} + L_{neg}$, where $L_{real}$ is the loss of real instances, $L_{gen}$ is the loss of generated instances, and $L_{neg}$ is the loss of negative instances.
Consequently, we can use $ \nabla_{\theta}{L_{gen}} $ to replace $ \nabla_{\theta} ( L_{\widehat{\mathcal{R}}_b}(\theta) - L_{\mathcal{R}_b}(\theta) )$

\section{More ablations}
\textbf{Momentum Coefficient $\beta$.}
In \cref{alg:batch3}, we introduce a momentum coefficient $\beta$ to update the grad cache. Here we explore the effect of different $\beta$ on the model performance.
A larger beta signifies a greater focus on global information, while a smaller beta indicates a higher attention to the current test batch $\mathcal{U}_b$.
Detailed results are presented in \cref{tab:beta}. Observations suggest that when $\beta$ is 0.1, the performance is the best, which is also the $\beta$ we finally adopted.

\begin{table}[h]
    \centering
    \caption{Comparison of different $\beta$ for updating grad cache.}
    \label{tab:beta}
    \begin{tabular}{@{}ccccc@{}}
\toprule
\multicolumn{1}{l}{$\beta$} & $AP^{box}$     & $AP^{mask}$    & $AP_r^{box}$   & $AP_r^{mask}$  \\ \midrule
0.05                        & 35.14          & 31.24          & 27.24          & 24.50          \\
0.10                        & \textbf{35.40} & \textbf{31.56} & \textbf{27.95} & \textbf{25.43} \\
0.30                        & 34.84          & 31.18          & 26.12          & 24.32          \\
0.50                        & 34.87          & 31.03          & 25.80          & 22.88          \\
0.80                        & 34.50          & 30.72          & 24.37          & 21.89          \\ \bottomrule
\end{tabular}%
\end{table}

\textbf{Contribution threshold $\tau$.}
In \cref{alg:batch3}, we incorporate a contribution threshold $\tau$, intended for filtering the produced data. Here we investigate the impact of varying values of   $\tau$ on the model's performance.
The larger $\tau$ implies a stricter filtration of the generated data, while the smaller $\tau$ signifies a looser filtering of the generated data.
The specific results are shown in \cref{tab:tau}. We can see the performance is optimal when $\tau$ equals -0.05, which is also the $\tau$ we eventually settle on for our final model.

\begin{table}[h]
    \centering
    \caption{Comparison of different $\tau$ for filtering generated data.}
    \label{tab:tau}
   
   \begin{tabular}{@{}rcccc@{}}
\toprule
\multicolumn{1}{l}{$\tau$} & \multicolumn{1}{c}{$AP^{box}$} & \multicolumn{1}{c}{$AP^{mask}$} & \multicolumn{1}{c}{$AP_r^{box}$} & \multicolumn{1}{c}{$AP_r^{mask}$} \\ \midrule
-0.10                      & 34.68                          & 30.80                           & 26.60                            & 24.40                             \\
-0.05                      & \textbf{35.40}                 & \textbf{31.56}                  & \textbf{27.95}                   & \textbf{25.43}                    \\
0.00                       & 34.72                          & 30.98                           & 24.96                            & 22.79                             \\
0.05                       & 34.29                          & 30.55                           & 23.75                            & 21.71                             \\ \bottomrule
\end{tabular}%
    
\end{table}

\begin{figure}
    \centering
    \includegraphics[width=1\linewidth]{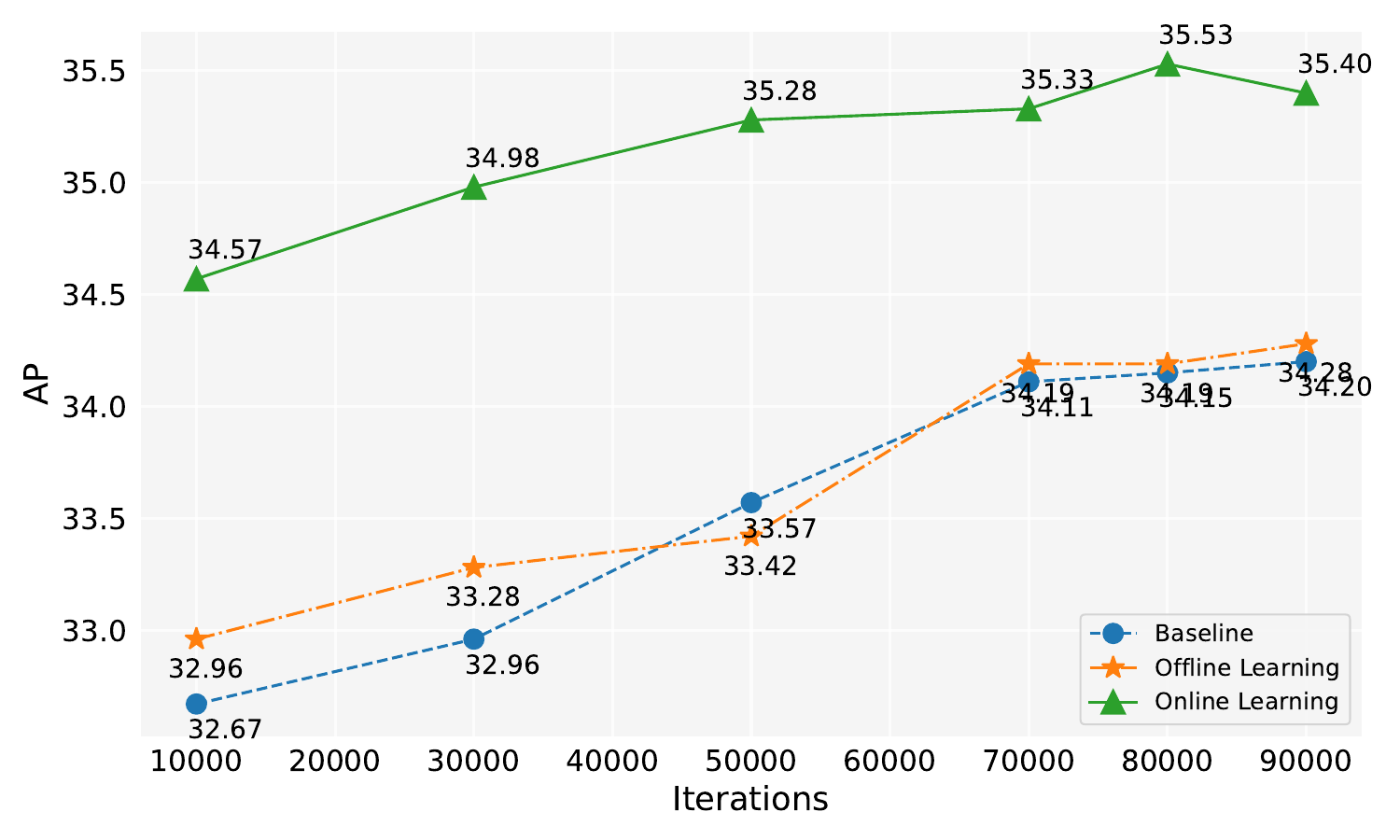}
    \caption{Performance of the model under different iterations.}
    \label{fig:all_3_iter}
\end{figure}

\textbf{Online learning vs.\  Offline learning}
We compare online learning and offline learning under different iterations. The result is shown in \cref{fig:all_3_iter}.

\section{Discussion}
\subsection{Comparing with existing methods}
\begin{table}[h]
\caption{Comparing with existing methods}
\label{tab:sample-table}
\begin{center}

\begin{tabular}{lcccccc}
\toprule
Method & Data Scale & Downstream & Cost & Quality & Domain Diff \\
\midrule
Traditional Active Learning & Limited & $\surd$ & High & $\surd$ & $\surd$ \\
Generated Data Filtering Methods & Unlimited & $\times$ & Low & $\times$ & $\times$ \\
Generative Active Learning & Unlimited & $\surd$ & Low & $\times$ & $\surd$ \\
\bottomrule
\end{tabular}

\end{center}
\vskip -0.1in
\end{table}
 We've drawn the \cref{tab:sample-table}, analyzing our setting compared to previous active learning or generative data filtration methods. We've conducted analysis from aspects of data scale, whether it's oriented towards downstream tasks, label quality, labeling costs, and whether there exists domain difference (between generated and real data).
\subsection{Analysis of the computational cost}
\begin{table}[h]
\centering
\caption{Analysis of the computational cost of different algorithms}
\begin{tabular}{ccc}
\toprule
\textbf{Methods} & \textbf{Total Time} & \textbf{Max Memory per GPU} \\
\midrule
Baseline & 17h & 6534M \\

Loss estimate (Algorithm 2) & 31h & 7114M \\

Grad cache (Algorithm 3) & 21h & 9836M \\
\bottomrule
\end{tabular}
\end{table}

We recorded the training duration and GPU memory usage for training 90,000 iterations with 4 Nvidia 4090 GPUs. It can be observed that our method based on Grad cache increases the GPU memory usage compared to Loss estimate, but it significantly reduces the training time. Compared with our Baseline, the additional time and memory overheads are within an acceptable range.

\subsection{Future work}
We hope that this paper can provide more inspiration to the academic community  on how to utilize generated data and how to design better data analysis methods.
It should be pointed out that our method is not limited to specific tasks or specific model architectures. In this work, for the convenience of comparison with the baseline, we use the same dataset and model architecture as the baseline.
We hope that in future work, we can further verify it on more tasks and model architectures. At the same time, we can also design more flexible and controllable evaluation functions to better utilize generated data.
For example, in this paper, when filtering the data with a gradient, there is a trade-off between diversity and consistency. For rare categories in the data, due to the small number of real data itself, diversity should be considered more, while for common categories, due to the large number of real data itself, consistency should be considered more. Therefore, in the future, we can consider adopting a dynamic strategy for different categories. In the long run, our current research is done under the premise of a fixed generative model. A more ideal situation is to involve the generative model in this loop, further optimizing the generative model based on the downstream model's feedback, to achieve a true ``generative model in the 
loop''.

\section{Visualization}
\subsection{Selected and Discarded Samples}
\begin{figure*}[h]
    \centering
    \includegraphics[width=\linewidth]{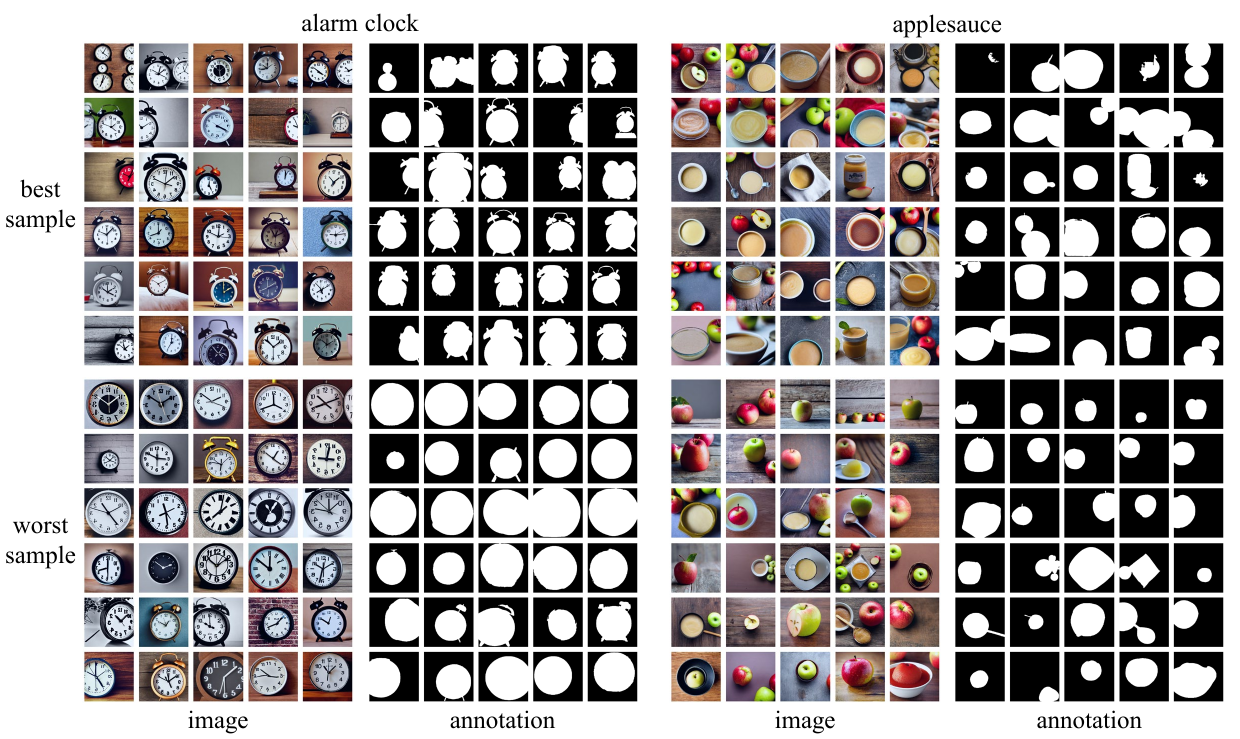}
    \caption{Examples of selected and discarded samples.}
    \label{fig:suppl_select}
\end{figure*}

We show some samples selected and discarded by our method in Figure~\ref{fig:suppl_select}. Our proposed method is able to select high-quality samples (best sample) while filtering out low-quality samples (worst sample), which can effectively improve the data learning efficiency of the model. 
For example, our method is capable of identifying accurately segmented data for applesauce. In cases where applesauce is not present in the generated raw image or is not encompassed within the segmentation mask, our method can discard such samples. For alarm clocks, our method tends to choose images with more complex appearances.

\subsection{Instance Augmentation}
\begin{figure*}
    \centering
    \includegraphics[width=0.9\linewidth]{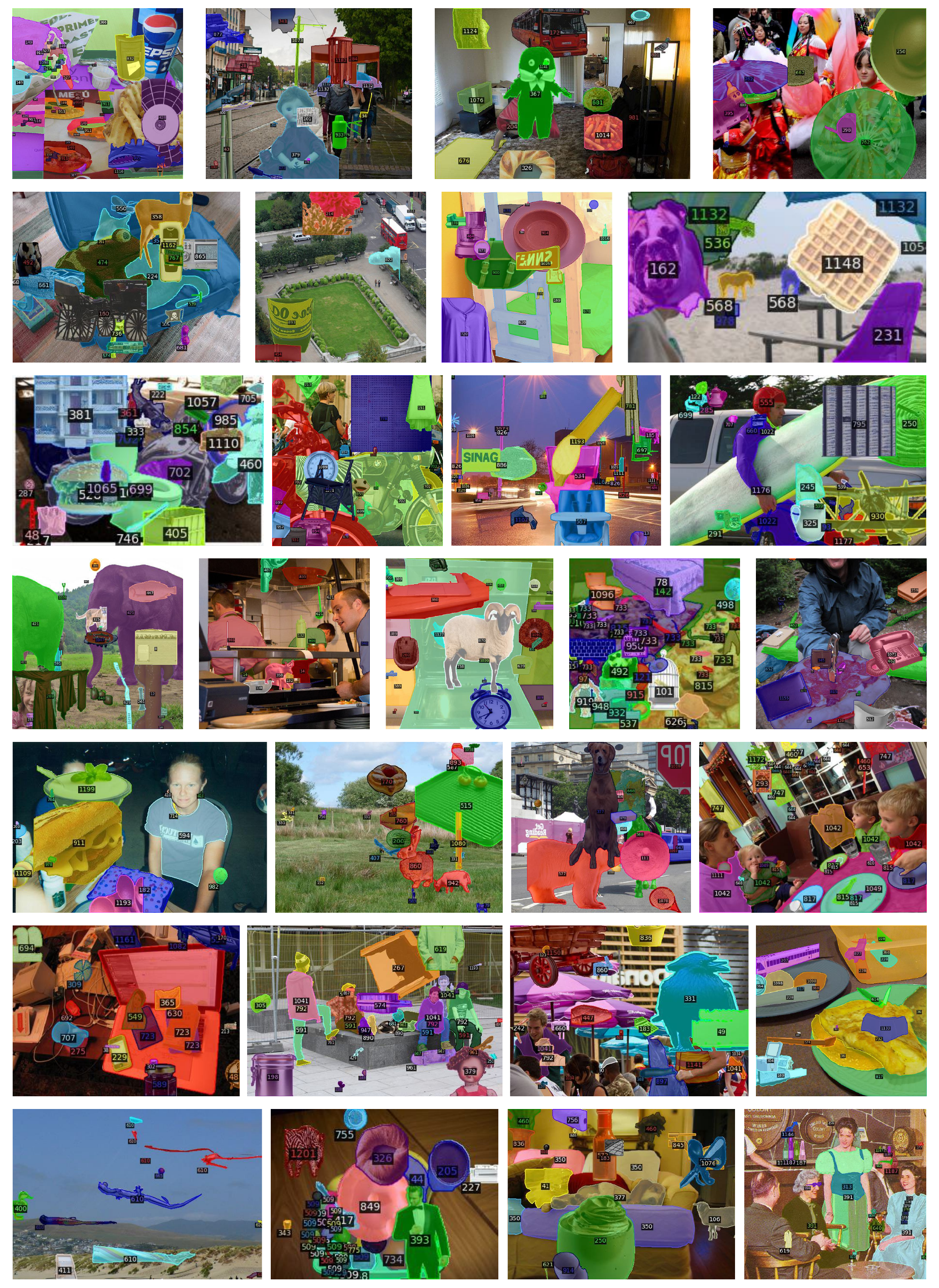}
    \caption{Examples of augmented data.}
    \label{fig:suppl_gen}
\end{figure*}

We present some augmented data in Figure~\ref{fig:suppl_gen}. By randomly pasting generated samples onto the LVIS training set, we effectively enrich the complexity of the scenes and thus increase the model's learning efficiency on the generated data.

\end{document}